\documentclass[10pt, a4paper]{article}
\usepackage{lrec-coling2024} 

\title{Empowering Small-Scale Knowledge Graphs: A Strategy of Leveraging General-Purpose Knowledge Graphs for Enriched Embeddings}
\name{Albert Sawczyn, Jakub Binkowski, Piotr Bielak, Tomasz Kajdanowicz} 

\address{Wrocław University of Science and Technology \\
         wybrzeże Stanisława Wyspiańskiego 27 \\
         albert.sawczyn@pwr.edu.pl \\}

\abstract{
Knowledge-intensive tasks pose a significant challenge for Machine Learning (ML) techniques. Commonly adopted methods, such as Large Language Models (LLMs), often exhibit limitations when applied to such tasks. Nevertheless, there have been notable endeavours to mitigate these challenges, with a significant emphasis on augmenting LLMs through Knowledge Graphs (KGs). While KGs provide many advantages for representing knowledge, their development costs can deter extensive research and applications. Addressing this limitation, we introduce a framework for enriching embeddings of small-scale \textit{domain-specific} Knowledge Graphs with well-established \textit{general-purpose} KGs. Adopting our method, a modest \textit{domain-specific} KG can benefit from a performance boost in downstream tasks when linked to a substantial \textit{general-purpose} KG. Experimental evaluations demonstrate a notable enhancement, with up to a $44\%$ increase observed in the $Hits@10$ metric. This relatively unexplored research direction can catalyze more frequent incorporation of KGs in knowledge-intensive tasks, resulting in more robust, reliable ML implementations, which hallucinates less than prevalent LLM solutions.
\\ \newline  }
\usepackage{amsmath}
\usepackage{adjustbox}
\usepackage{siunitx}
\usepackage[textsize=tiny]{todonotes}
\usepackage{xstring}
\usepackage{color}
\usepackage{graphicx}
\usepackage{tabularx}
\usepackage{titlesec}
\usepackage{amssymb}
\usepackage{float}
\usepackage{placeins}

\newcommand\Tstrut{\rule{0pt}{1.9ex}}         
  
\begin{document}

\maketitleabstract

\section{Introduction} \label{sec:intro}

In recent years, machine learning has started to be commonly used in numerous applications. Recent advancements in Natural Language Processing, especially the emergence of Large Language Models \cite{devlinBERTPretrainingDeep2019, brown2020language, touvron2023llama}, have sparked the adoption of AI-based solutions in many real-world circumstances. The ever-growing demand for automating knowledge-intensive tasks \cite{petroniKILTBenchmarkKnowledge2021} motivates many researchers to utilize NLP solutions, but these are prone to errors \cite{adlakha2023evaluating, bowman2023things}. Knowledge-intensive tasks are characterized by the need for comprehensive understanding and reasoning, meaning they require AI models to go beyond surface-level data and tap into a wealth of interconnected information, much like a human expert would \cite{lewisRetrievalAugmentedGenerationKnowledgeIntensive2020}, while also accounting for the frequent updates of its knowledge. Examples of knowledge-intensive tasks include question-answering systems, fact-checking, information retrieval, and recommendation systems \cite{petroni2019language}. To deal with the peculiarities of such a setting, one might leverage Knowledge Graphs (KGs) that hold knowledge specific to the domain of interest. That is why there were several attempts to combine Language Models with KGs \cite{colonhernandez2021combining}. However, building comprehensive KGs is an expensive and complex task \cite{Sequeda2021}, thus often avoided, which might hinder the development of many endeavours. Therefore, to foster the further development of KG-based systems, we propose a framework that combines well-developed KGs with smaller and still-developing domain-specific KGs. 

\begin{figure}[ht!]
    \centering
    \includegraphics[width=0.95\linewidth]{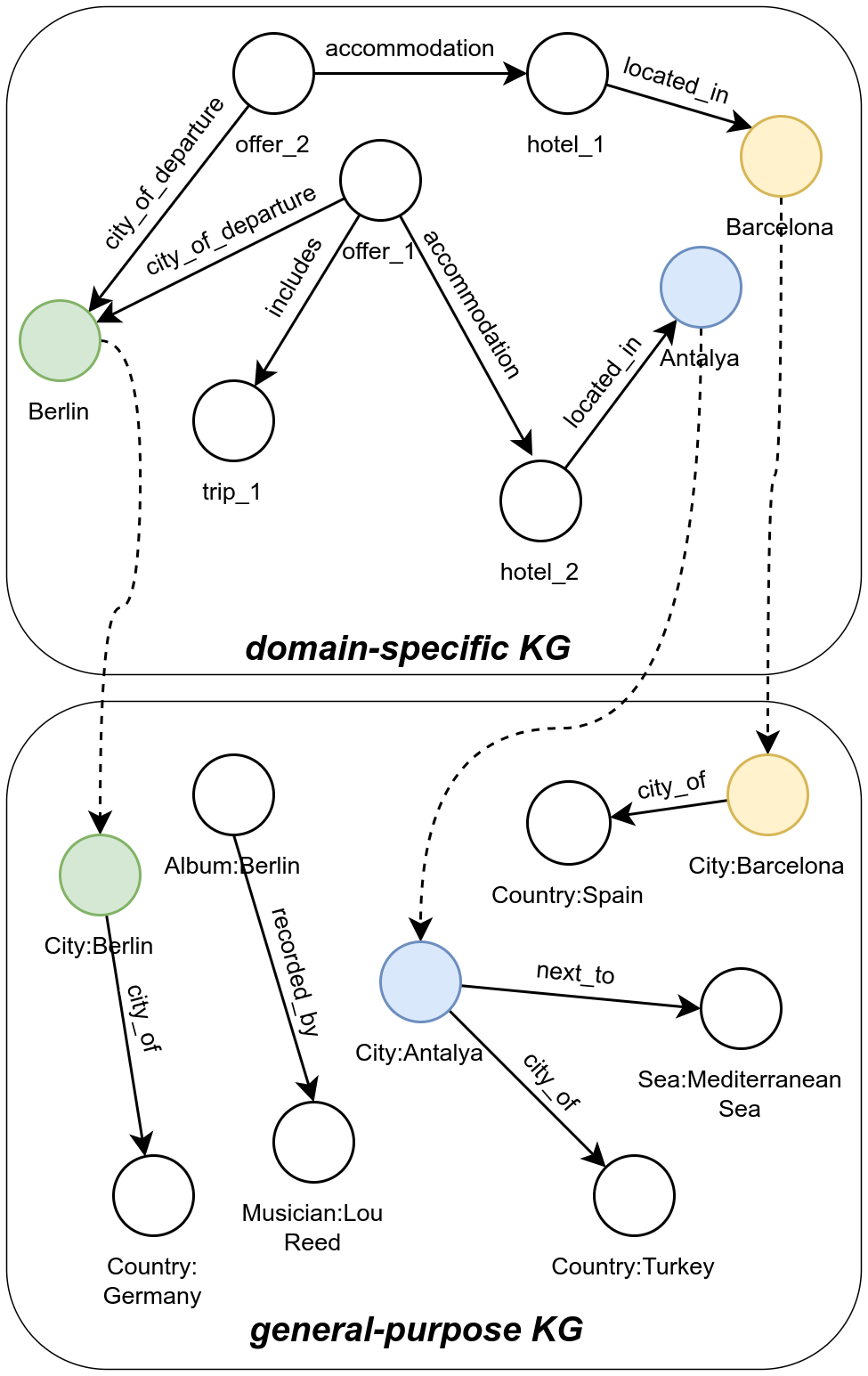}
    \caption{Diagram presenting two aligned and linked Knowledge Graphs: \textit{domain-specific} (upper), \textit{general-purpose} (bottom). Artificial links, marked with dashed lines, connect two KG.}
    \label{fig:alignment}
\end{figure}

Creating a KG is a resource-intensive process that requires a significant investment of time and capital. This makes it particularly challenging for small to medium-sized teams, startups, or groups in academia, who often face resource constraints \cite{jiSurveyKnowledgeGraphs2022}. A small, developing KG might be insufficient as such a graph often suffers from limited relational structures, sparse entity interactions, and reduced contextual information. This scarcity of data can limit the model's ability to generalize well, capture intricate semantic relationships, or make accurate predictions. To leverage the full potential of Knowledge Graphs and enable their early adoption, we propose an unexplored approach -- to focus on creating a smaller, \textit{domain-specific KG (DKG)}, which addresses specific needs, and combining it with a broader but well-developed \textit{general-purpose KG (GKG)}. This enables small KGs to immediately leverage additional knowledge in downstream tasks, eventually leading to better performance. This concept is visualized in Figure \ref{fig:alignment}, i.e., we automatically link \textit{domain-specific} KG to \textit{general-purpose} KG using \textit{entity alignment} (or \textit{alignment} for short)  and \textit{linking} operations, which combine two KGs by matching their nodes and putting artificial links between them. Further, we train a translation-based embedding method in a KG completion task. We demonstrate that training such a model on the linked KGs can enhance performance in a downstream task of KG completion on the \textit{domain-specific} graph compared to a scenario without \textit{general-purpose} KG.

We summarize our contributions as follows:
\begin{enumerate}
    \item We propose a generic and modular framework for enriching small and \textit{domain-specific} KGs with \textit{general-purpose} KGs using \textit{alignment} and \textit{linking} operation. The framework is tailored to real-world scenarios. 
    \item We propose an evaluation methodology for experiments and conduct empirical studies on synthetic and real-world scenarios. In both settings, evaluation is conducted under rigorous conditions, i.e., DKG lacks many nodes and links to perform well on a downstream task.
    \item We propose a weighted loss function to mitigate entity alignment's negative effects between two considered KGs.
\end{enumerate}

The work adheres to the widely-renowned reproducibility standards. We publish full implementation and data (datasets, models) in the GitHub repository\footnote{\url{https://github.com/graphml-lab-pwr/empowering-small-scale-kg}}, and experiment tracking logs on Weights\&Biases \cite{wandb}\footnote{\url{https://wandb.ai/graph-ml-lab-wust/empowering-small-scale-kg}}.

\section{Related Work}
To leverage a Knowledge Graph in a downstream task, like KG completion, one has to obtain representations \cite{bengioRepresentationLearningReview2013} of its entities (and relations), which are feature vectors describing these objects \cite{hamiltonRepresentationLearningGraphs2017}, similar to (contextual) word embeddings in Natural Language Processing. This pertains to using standalone KG or fusing it with Large Language Model \cite{colonhernandez2021combining}. Recently, learning representation on graphs, including Knowledge Graphs, gained much interest in the community, primarily due to the development of Graph Neural Networks \cite{gilmerNeuralMessagePassing2017, kipfSemisupervisedClassificationGraph2017}. Representation learning of KGs aims to map entities and relationships into low-dimensional vector spaces, thereby indirectly capturing their semantic meaning \cite{bengioRepresentationLearningReview2013}. However, unlike the most prevalent setup of processing homogeneous graphs, learning representations on KGs requires tackling two main challenges. First, the vast size of KGs poses issues as most established methods lose efficacy when handling billions of data points. Second, KGs often lack completeness due to inherent imperfections in their creation process \cite{pujaraSparsityNoiseWhere2017}. 

Nonetheless, several groups of representation learning methods for KGs can be found in the literature. Among these groups, translation-based, GNN-based \cite{zhangBenchmarkComprehensiveSurvey2022a}, and factorization-based \cite{trouillon2017knowledge, Balazevic_2019} methods stand out the most. The TransE model \cite{bordesTranslatingEmbeddingsModeling2013} is one of the most common translation-based models in KG representation learning. It operates on the principle of representing relations as translations in the embedding space, i.e., the embedding of the edge should move the embedding of the head entity to end up with an embedding of the tail entity. Due to several limitations of TransE, models such as RotatE \cite{sun2018rotate}, BoxE \cite{abboudBoxEBoxEmbedding2020}, and QuatE \cite{zhangQuaternionKnowledgeGraph2019} were proposed. For instance, RotatE leveraged complex space, and the representation of each relation is a rotation from head to tail entity. Thus, RotatE can encapsulate information about symmetry, anti-symmetry, inversion, and composition in the learned representations \cite{sun2018rotate}.

The requirements of KGs for learning the attribute features and structural features of entities and relationships can be met by GNN-based methods, which integrate the topology and attribute information. An early yet still prevalent approach is the Relational Graph Convolutional Networks (R-GCN) \cite{schlichtkrullModelingRelationalData2018}, a relation-aware variant of GCN. Later significant methods include 
CompGCN \cite{vashishthCompositionbasedMultiRelationalGraph2019}, KBGAT \cite{nathaniLearningAttentionbasedEmbeddings2019}. While GNNs showed promising performance in several studies, we chose the family of translation-based methods for learning representations on combined graphs due to their comparable performance to GNN models with higher stability and shorter learning times \cite{aliBringingLight2022}.

Besides, in our work, we propose combining two graphs by entity alignment and linking operation before learning representations of entities. Given two KGs, alignment aims to automatically find matching pairs of entities corresponding to the same real-world entity \cite{zhangBenchmarkComprehensiveSurvey2022a}, and linking creates artificial connections between aligned entities. While several approaches for data-driven alignment \cite{zengComprehensiveSurveyEntity2021} were proposed, including MTransE \cite{chenMultilingualKnowledgeGraph2017}, bootstrapping \cite{bordesTranslatingEmbeddingsModeling2013}, or AttrE \cite{trisedyaEntityAlignmentKnowledge2019},
current benchmarks for assessing these methods exhibit notable restrictions: assumptions of bijection, a lack of name variety, and small scale of graphs. These limitations starkly contrast with the realities encountered in real-world applications, which tend to be larger in scale, more diverse, and far from a 1:1 mapping \cite{zhangBenchmarkComprehensiveSurvey2022a}. In our work, we accept that alignment methods might be imperfect and link two graphs based on simple similarity scores between entities. Instead of optimizing loss terms related to alignment, as done in other studies \cite{selfKG, JAPE}, we directly optimize KG completion loss over linked KGs. The previous literature tackled graph linking, but not our approach for embedding enrichment. While aligned with the semantic web vision \cite{gruber1995toward, berners1998semantic}, our focus goes beyond traditional discourse. Hence, to the best of our knowledge, this is the first investigation of this type.

\begin{figure*}[ht!]
    \centering
    \includegraphics[width=\textwidth]{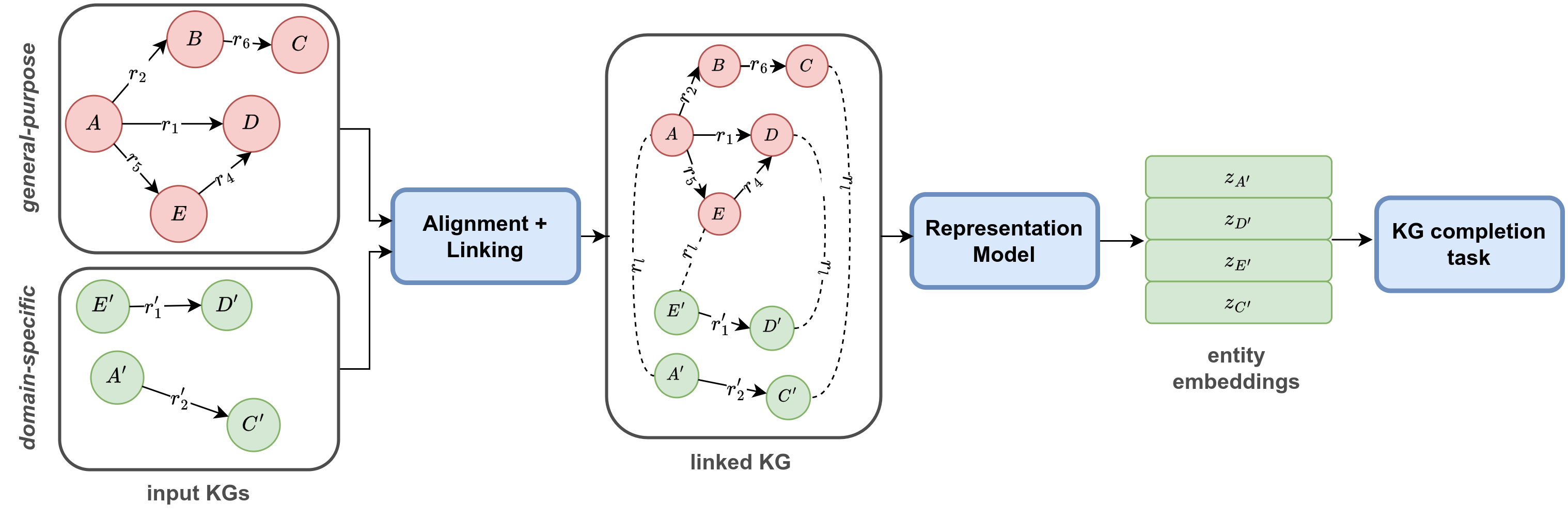}
    \caption{Overview diagram of the proposed framework's pipeline.}
    \label{fig:overview}
\end{figure*}

\section{Methodology}
In this work, we propose learning Knowledge Graph completion in a small-scale domain-specific KG by connecting it with a larger general-purpose KG by alignment and linking operation. Further, such linked KG is fed to a downstream task model, i.e., KG completion. The entire framework is schematically presented in Figure \ref{fig:overview}, and below, we formally introduce each component.

\subsection{Notation}
We denote Knowledge Graph as a multi-relational heterogeneous graph $\mathcal{G}=(\mathcal{E}, \mathcal{R}, \mathcal{T})$, consisting of a set of entities $\mathcal{E}$, a set of relation predicates $\mathcal{R}$, and a set of triples (facts) $\mathcal{T}$. A single triple $(h, r, t) \in \mathcal{T}$ indicates a relation predicate $r$ between two entities, a head entity $h$ and a tail entity $t$, where $h, t \in \mathcal{E}$ and $r \in \mathcal{R}$ \cite{hamiltonRepresentationLearningGraphs2017}.

\subsection{Alignment and Linking} \label{sec:alignment_linking}
First, we define \textit{entity alignment} in the context of two KGs: a domain-specific $(d)$ one and a general-purpose $(g)$ one: 
\begin{align*}
    \mathcal{G}_d=(\mathcal{E}_d, \mathcal{R}_d, \mathcal{T}_d), \;
    \mathcal{G}_g=(\mathcal{E}_g, \mathcal{R}_g, \mathcal{T}_g)
\end{align*}
\textit{Alignment} aims to discover entity pairs $(e^{(d)}_i, e^{(g)}_j)$, $e^{(d)}_i \in \mathcal{E}_d$, $e^{(g)}_j \in \mathcal{E}_g$ where $e^{(d)}_i$ and $e^{(g)}_j$ correspond to the same real-world entity. In other words, $e^{(d)}_i$ and $e^{(g)}_j$ are aligned entities \cite{zengComprehensiveSurveyEntity2021}. We propose a simple strategy for alignment, detailed below.

We observe that each entity has assigned a word label, such as "car", and we can vectorize it using a text embedding method of our preference. Further, to achieve the representation $x(e_i)$ for each entity $e_i$, we concatenate three terms: embedding of entity label, mean embedding of outgoing neighbours' labels and mean embedding of ingoing neighbours' labels:
\begin{equation*}
    x(e_i) = \left[ \phi(e_i) \, \Bigg\Vert \, \sum_{e_u \in \mathcal{O}(e_i)} \frac{\phi(e_u)}{|\mathcal{O}(e_i)|} \, \Bigg\Vert \, \sum_{e_v \in \mathcal{I}(e_i)} \frac{\phi(e_v)}{|\mathcal{I}(e_i)|} \right],
\end{equation*}
where embedding $\phi(\cdot)$ is a pre-trained text embedding, $\Big\Vert$ denotes the concatenation operator, $\mathcal{O}(e_i)$ returns outgoing nodes and $\mathcal{I}(e_i)$ returns ingoing nodes.

Then, for each entity in the DKG, we find $k$ nearest entities in the GKG, based on $x(\cdot)$ embeddings, where $k$ is a hyperparameter. The nearest neighbour search leverages a vector distance metric, e.g., euclidean, cosine. Further, for any pair of found neighbours $e^{(d)}_i$ and $e^{(g)}_j$, we create an artificial triple $(e^{(d)}_i, r_l, e^{(g)}_j) \in \mathcal{T}_l$ between two KGs and call it \textit{linking}. The linked KG is formalized as:
\begin{align*}
    \mathcal{G}_{l} = (\mathcal{E}_d \cup \mathcal{E}_g, \mathcal{R}_d \cup \mathcal{R}_g \cup \{r_l\}, \mathcal{T}_d \cup \mathcal{T}_g \cup \mathcal{T}_l ),
\end{align*}

\subsection{Representation Model}
Representation learning aims to learn a model that extracts relevant and semantically rich features from the underlying data. In the context of knowledge graphs, we formalize this task as learning (parametrized) function $f_\theta: \mathcal{G}_l \rightarrow Z$, where $Z \in \mathbf{R}^{N \times d}$ is a matrix of entity representations, where $N$ is the number of entities, and $d$ is a dimensionality of the representation. In other words, we expect to optimize $\theta$ under a particular loss function. It is worth noting that Representation Model can be trained end-to-end with KG completion in a supervised setting \cite{hamiltonRepresentationLearningGraphs2017}, which is also our case. We present the particular choice for the representation model in Section \ref{sec:experimental_setup}.

\subsection{KG completion and Weighted Loss function}
In the KG completion task, for every pair of entities in KG, the model predicts the existence of each possible relation between them \cite{hamiltonRepresentationLearningGraphs2017}. During training, the model treats existing relations as positive examples, and to avoid collapse, negative relations are sampled (non-existing ones). In our setting, we assume that the alignment is imperfect. Hence, we introduce a loss function which differently weights relations arising from the linking operation by the distance score used in nearest neighbour search, as shown in Equation \eqref{eq:loss}:

\begin{equation}
\label{eq:loss}
    \mathcal{L}(\mathcal{T}, \tilde{\mathcal{T}}; \theta) = \sum_{t \in \mathcal{T}_d \cup \mathcal{T}_g} L(t, \tilde{\mathcal{T}}; \theta) + \sum_{s \in \mathcal{T}_l} w_s \cdot L(s, \tilde{\mathcal{T}}; \theta),
\end{equation}

where $\mathcal{L}(\cdot)$ denotes loss function for the entire model, $L(\cdot)$ is a loss function of the Representation Model, and $\tilde{\mathcal{T}}$ represents the sampled negative relations, and $w_s = 1 / (1 + distance(x(e_i), x(e_j)))$ represents similarity score between representation of aligned entities $e_i$ and $e_j$.

\section{Experiments}
We proposed a framework that performs effectively in real situations, mainly when DKG are much smaller than GKG. Since existing datasets do not capture this scenario, we designed a custom evaluation procedure to fit this context. In particular, two distinct scenarios were implemented: synthetic and real-world. 

\subsection{Datasets}
In the experiments, we leveraged common KG datasets which emulate the realistic conditions of size diversity, domain overlapping, and mapping possibilities. \textbf{WN18RR} \citeplanguageresource{dettmersConvolutional2DKnowledge2018} -- derived from WordNet \citeplanguageresource{miller1995wordnet}, a lexical database of English focuses on semantic relations between words. \textbf{FB15k-237} \citeplanguageresource{toutanovaObservedLatentFeatures2015} -- sourced from Freebase, a knowledge graph that spans encyclopedic topics. \textbf{WD50K} \citeplanguageresource{galkinMessagePassingHyperrelational2020} -- an encyclopedic KG extracted from Wikidata \citeplanguageresource{vrandevcic2014wikidata} based on the seed nodes corresponding to entities from FB15k-237. \textbf{ConceptNet} \citeplanguageresource{speerConceptNetOpenMultilingual2017} -- semantic network encapsulating commonsense knowledge (we used version 5.7.0 and extracted the English segment of this KG). \textbf{YAGO3-10} \citeplanguageresource{mahdisoltaniYAGO3KnowledgeBase2015} -- a commonsense KG sourced from multiples sources, including WordNet \citeplanguageresource{miller1995wordnet}.

To access textual labels of entities, we used decoded versions of WN18RR, FB15k-237, and YAGO3-10, which are available in the repository\footnote{\url{https://github.com/villmow/datasets_knowledge_embedding}}. For WD50K, we utilized Wikidata API\footnote{\url{https://www.wikidata.org/wiki/Wikidata:REST_API}} to map entity IDs to their textual labels. ConceptNet inherently provides textual labels for its entities, eliminating the need for decoding. Finally, we removed all terms used for identity identification, e.g., \texttt{absorb.v.01} $\rightarrow$ \texttt{absorb}. The detailed statistics of the used datasets are presented in Appendix \ref{sec:supplementary_stats}: Table \ref{tab:original-stats}.

\subsection{Evaluation procedure} 
An approach we adopted involves taking a dataset and then sampling it for use as the DKG that we used for both synthetic and real-world scenarios we describe below.

\subsubsection{Sampling} \label{sec:sampling}

To simulate the early stages of KG development, we introduce three distinct graph sampling strategies:

\textbf{Triple sampling} draws triples (facts) at random. While this directly affects the number of facts in the sampled KG, it retains all entities and relations in the chosen triples. However, this might result in some entities or relations being disconnected.

\textbf{Node sampling} selects entities and all triples associated with it. This approach often produces a more cohesive subgraph but may exclude distant interactions.

\textbf{Relation sampling} selects relations and all triples containing this relation. This method focuses on preserving specific interactions, potentially leading to diverse entities included, but only via the selected relations.

Each strategy is parameterized by the probability $p$ of keeping a triple, node, or relation in the sampled graph, thus determining the size of the resulting DKG. It offers precise control over the sampling intensity, allowing us to craft KGs of varying sizes and complexities to mimic different stages in developing a KG. The detailed statistics of the sampled datasets are presented in Appendix \ref{sec:supplementary_stats}: Table \ref{tab:sampled-stats}.

Given the standard transductive setting, we added an initial step to the sampling procedure to ensure that all entities and relations from the validation (val) and testing (test) subsets are preserved in a sampled dataset. This allowed to compare across varying sampling parameters. For triple sampling, this entails having at least one triple for every entity and relation from the val/test. In node sampling, for every entity from these subsets, we ensure the preservation of at least one connected entity, guaranteeing the inclusion of at least one associated triple. However, depending on the value of $p$, retaining all val/test data might be impossible, leading to varying minimum $p$ values across datasets. Notably, in relation sampling, retaining the val/test subsets is impossible. Due to that, performance metrics derived from these sampled subsets should be interpreted cautiously, avoiding direct comparisons among themselves and with other datasets.

\subsubsection{Scenarios}

\paragraph{Synthetic scenario} In this scenario, the GKG is the original version of the graph, while the sampled version is taken as DKG. Primarily, the synthetic scenario provides a controlled environment that mimics realistic conditions of non-proportional graphs. It guarantees that the knowledge domains of the two KGs overlap perfectly, ensuring that a direct 1:1 alignment is feasible. This controlled alignment, in turn, provides a more straightforward context to evaluate and understand the properties of our proposed framework.

\paragraph{Real-world scenario} Here, an external, distinct KG takes on the role of the GKG, i.e., the two KGs are inherently different. This mirrors realistic scenarios where it is needed to enhance the efficacy of an underdeveloped knowledge graph by enriching it with a more extensive and well-established KG. For instance, consider the sampled WN18RR linked to ConceptNet, which encompasses knowledge spanning the domain of WordNet. Unlike the synthetic scenario, this real-world setup does not offer the same level of control for the experimental phase.

\subsection{Experimental setup} \label{sec:experimental_setup}

\subsubsection{Model setup} \label{sec:model_setup}

To perform alignment, we selected fastText word embeddings \cite{joulin2016bag}, as these turned out to be effective in the KG domain \cite{selfKG}. During linking, we performed $k$ nearest neighbours search based on the Euclidean distance, leveraging the FAISS library \cite{johnson2019billion}. Further, as a Representation Model, we utilized RotatE \cite{sun2018rotate}, which optimizes lookup embeddings for each entity and relation. We trained RotatE end-to-end with the KG completion task (without updating FastText embeddings). We considered two variants of the loss function: the first setting involves the weighted loss function from Eq. \eqref{eq:loss} setting both $L(\cdot)$ terms to self-adversarial negative sampling loss \cite{sun2018rotate}; the second uses standard loss that is equivalent to setting all weights $w = 1.0$ in the weighted loss. To optimize the model, we leveraged the Adam optimizer \cite{kingma2014adam}. 

Based on the insights from  \citealp{aliBringingLight2022} combined with our empirical observations, the parameters were established as follows: the embedding dimension was set at $256$, the batch size at $512$, the learning rate at $0.004$, the margin at $9$, the adversarial temperature at $0.34$, for every positive $33$ negatives were generated, and the models were trained for $200$ epochs. 

During the synthetic scenario, the parameter $k$ was searched across values $\{1, 2, 3\}$. Based on the results of this exploration, $k$ was set to $1$, as it emerged as the most promising value for a real-world scenario. 
Additionally, due to the large size of GKGs in the real-world scenario, we considered cropping them to include only aligned entities and their $c$-hop neighborhood. This approach reduced experiment time and, as our empirical results showed, enhanced performance. 
Based on preliminary results, we defined specific cropping strategies ($c=\infty$ meaning not cropping) for each (DKG, GKG) pair, and then searched to select the best-performing one. These strategies are as follows: for (WN18RR, FB15k-237) $c=\infty$; for (FB15k-237, YAGO3-10), $c \in \{1, 2\}$; and for (WD50K, FB15k-237), we set $c \in \{1, \infty \}$. For all other pairs, we set $c=1$.

\subsubsection{Training \& evaluation}

DKGs were created by sampling based on strategies described in Section \ref{sec:sampling} with values $p \in \{0.4, 0.6, 0.8\}$. There were three types of training setups: \textit{single graph} (using only the DKG without any links to external graph); \textit{synthetic scenario} (linked with original GKG); \textit{real-world scenario} (linked with distinct GKG). Due to the non-trivial nature of GKG selection in the real-world scenario, we tested two different options for the GKG. DKGs were sampled from WN18RR, FB15k-237, and WD50K. For FB15k-237 and WD50K, node sampling had to be restricted due to the high diversity of nodes in val/test (see Section \ref{sec:sampling}). In the real-world scenario, the ConceptNet, FB15k-237, and WD50K were utilized as GKGs. Experiments conducted on a single graph or for a synthetic scenario were executed $3$ times each with a distinct random seed for both the sampling procedure and the model initialization. However, due to the low standard deviation observed across the runs, we reduced the number of runs to one for the real-world scenario. We reported the mean and standard deviation values of the runs. We used standard metrics for the KG completion task, computed on DKG: $Hits@10$,  Mean Rank ($MR$) and Mean Reciprocal Rank ($MRR$). During the hyperparameter search, the mean $Hits@10$ on val was chosen as the benchmark to determine the optimal set of hyperparameters.

The implementation used PyKEEN \cite{aliPyKEENPythonLibrary2021} and PyTorch \cite{paszkePyTorchImperativeStyle2019} libraries. To maintain reproducibility, we utilized DVC \cite{dvc} and Weights\&Biases \cite{wandb}.

\section{Results}

\subsection{Synthetic scenario} \label{sec:results-synthetic}
\begin{figure}[htb]
    \centering
    \includegraphics[width=\linewidth]{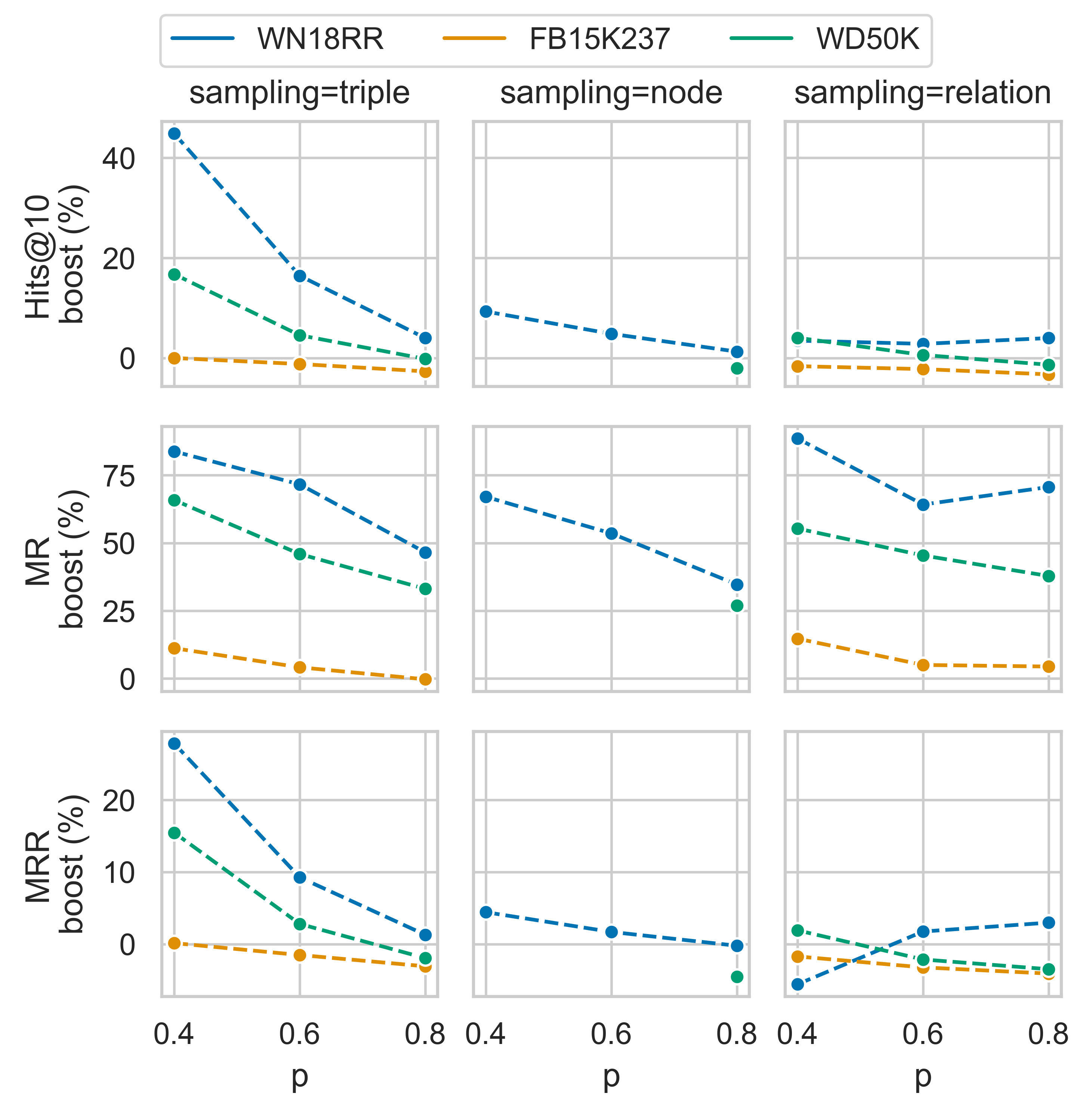}
    \caption{Synthetic scenario: performance boost across different sampling strategies and varying rates $p$. The boost is the performance improvement the framework achieves over a single graph. Please note that the relation sampling datasets should not be directly compared amongst themselves as the test was not preserved, causing high standard deviation (see \ref{sec:sampling}).}
    \label{fig:boost}
\end{figure}

Results from the synthetic scenario are presented in Table \ref{tab:results-synthetic}. For each combination of (dataset, sampling, $p$), we recorded the efficacy of the trained model on a \textit{single} graph and using the proposed framework (linked). The relative difference between them (\textit{boost}) is a crucial metric that evaluates the framework. The results show that we can obtain significant improvement by linking GKG. Nevertheless, the boost is strictly dependent on the sampling setting. The more rigorous the sampling setting, the greater the increase in effectiveness (see Figure \ref{fig:boost}). The negative correlation between $p$ and performance boost indicates that our framework can be helpful, especially in the harsh conditions of early stages of KG development or when it has limited data. Having only $40\%$ of triples we achieved $44.9\%/0.0\%/16.7\%$ $Hits@10$ boost on WN18RR/FB15k-237/WD50K respectively. However, the effectiveness differs among the datasets. For instance, while WN18RR experienced a significant boost, FB15k-237 showed a smaller increase. This variability can be attributed to each dataset's inherent characteristics and complexities. 

\bgroup

\begin{table*}[htb!]
  \centering
  \adjustbox{width=\linewidth}{%
    \sisetup{
      separate-uncertainty=true,
      multi-part-units=single,
      table-align-uncertainty=true,
      detect-weight=true,
     retain-zero-uncertainty=true
    }
    \setlength{\tabcolsep}{3pt}

    \begin{tabular}{
      |l
      S[table-format=1.1]|
      S[table-format=1.3(3)]
      S[table-format=0.3(3)]
      S[table-format=-1.1]|
      S[table-format=4.0(3)]
      S[table-format=4.0(3)]
      S[table-format=-1.1]|
      S[table-format=1.4(3)]
      S[table-format=1.4(3)]
      S[table-format=-1.1]|
    }
    \hline
       &  & \multicolumn{3}{c|}{$Hits@10\uparrow$} & \multicolumn{3}{c|}{$MR\downarrow$} & \multicolumn{3}{c|}{$MRR\uparrow$} \Tstrut\\
       {sampling} & {p} & {single} & {linked} & {boost(\%)} & {single} & {linked} & {boost(\%)} & {single} & {linked} & {boost(\%)} \\
       \hline \hline
 \multicolumn{11}{|c|}{\textbf{WN18RR}} \Tstrut \\
 \hline
triple & 0.4 & 0.347\pm0.005 & \bfseries 0.502\pm0.006 & 44.9 & 7681\pm 62 & \bfseries 1245\pm 44 & 83.8 & 
0.270\pm0.003 & \bfseries 0.345\pm0.001 & 27.8 \Tstrut \\
triple & 0.6 & 0.446\pm0.001 & \bfseries 0.519\pm0.004 & 16.4 & 4908\pm172 & \bfseries 1392\pm 15 & 71.6 & 
0.342\pm0.003 & \bfseries 0.373\pm0.004 & 9.3 \\
triple & 0.8 & 0.525\pm0.004 & \bfseries 0.546\pm0.004 & 4.0 & 2685\pm155 & \bfseries 1435\pm 52 & 46.6 & 
0.416\pm0.005 & \bfseries 0.421\pm0.003 & 1.3 \\
node & 0.4 & 0.546\pm0.004 & \bfseries 0.597\pm0.002 & 9.3 & 2164\pm 52 & \bfseries 713\pm 18 & 67.0 & 
0.473\pm0.002 & \bfseries 0.494\pm0.001 & 4.4 \\
node & 0.6 & 0.562\pm0.000 & \bfseries 0.590\pm0.001 & 4.8 & 2044\pm 29 & \bfseries 950\pm 15 & 53.6 & 
0.480\pm0.001 & \bfseries 0.488\pm0.001 & 1.7 \\
node & 0.8 & 0.576\pm0.004 & \bfseries 0.583\pm0.001 & 1.3 & 1829\pm  8 & \bfseries 1194\pm 20 & 34.7 & \bfseries 
0.484\pm0.001 & 0.482\pm0.001 & -0.2 \\
relation & 0.4 & 0.696\pm0.182 & \bfseries 0.721\pm0.184 & 3.5 & 1083\pm450 & \bfseries 124\pm 70 & 88.5 & 
\bfseries 0.590\pm0.233 & 0.557\pm0.212 & -5.6 \\
relation & 0.6 & 0.755\pm0.129 & \bfseries 0.777\pm0.143 & 2.8 & 1542\pm622 & \bfseries 552\pm649 & 64.2 & 
0.684\pm0.145 & \bfseries 0.696\pm0.160 & 1.7 \\
relation & 0.8 & 0.731\pm0.115 & \bfseries 0.760\pm0.135 & 4.0 & 2013\pm157 & \bfseries 591\pm671 & 70.7 & 
0.659\pm0.130 & \bfseries 0.678\pm0.149 & 3.0 \\
\hline
\multicolumn{2}{|c|}{Max boost (\%)}  & &  & 44.9 &  &  & 88.5 &  &  & 27.8 \Tstrut \\
\multicolumn{2}{|c|}{Mean boost (\%)}  & &  & 10.1 &  &  & 64.5 &  &  & 4.8 \\
\hline \hline
 \multicolumn{11}{|c|}{\textbf{FB15k-237}} \Tstrut  \\ 
 \hline 
triple & 0.4 & \bfseries 0.360\pm0.002 & 0.360\pm0.000 & -0.0 & 315\pm  4 & \bfseries 279\pm  1 & 11.2 & 
0.204\pm0.001 & \bfseries 0.204\pm0.001 & 0.1 \Tstrut \\
triple & 0.6 & \bfseries 0.398\pm0.001 & 0.393\pm0.003 & -1.2 & 242\pm  1 & \bfseries 231\pm  0 & 4.2 & \bfseries 
0.227\pm0.001 & 0.224\pm0.002 & -1.5 \\
triple & 0.8 & \bfseries 0.442\pm0.000 & 0.430\pm0.002 & -2.7 & \bfseries 194\pm  2 & 195\pm  1 & -0.3 & \bfseries 
0.254\pm0.001 & 0.246\pm0.000 & -3.1 \\
relation & 0.4 & \bfseries 0.480\pm0.009 & 0.472\pm0.010 & -1.6 & 256\pm 68 & \bfseries 218\pm 61 & 14.7 & 
\bfseries 0.297\pm0.019 & 0.292\pm0.018 & -1.7 \\
relation & 0.6 & \bfseries 0.490\pm0.009 & 0.479\pm0.009 & -2.2 & 197\pm 39 & \bfseries 187\pm 39 & 5.0 & \bfseries
0.308\pm0.002 & 0.298\pm0.000 & -3.2 \\
relation & 0.8 & \bfseries 0.502\pm0.014 & 0.485\pm0.011 & -3.3 & 169\pm 17 & \bfseries 162\pm 12 & 4.4 & \bfseries
0.304\pm0.005 & 0.291\pm0.004 & -4.1 \\
\hline
\multicolumn{2}{|c|}{Max boost (\%)} &  &  & 0.0 &  &  & 14.7 &  &  & 0.1 \Tstrut\\
 \multicolumn{2}{|c|}{Mean boost (\%)}  &  &  & -1.8 &  &  & 6.5 &  &  & -2.3 \\
\hline\hline
 \multicolumn{11}{|c|}{\textbf{WD50K}}  \Tstrut \\
 \hline
triple & 0.4 & 0.283\pm0.001 & \bfseries 0.330\pm0.000 & 16.7 & 1713\pm 42 & \bfseries 585\pm  3 & 65.8 & 
0.164\pm0.001 & \bfseries 0.189\pm0.001 & 15.5 \Tstrut \\
triple & 0.6 & 0.353\pm0.000 & \bfseries 0.369\pm0.001 & 4.5 & 931\pm 20 & \bfseries 503\pm  5 & 45.9 & 
0.208\pm0.001 & \bfseries 0.213\pm0.000 & 2.8 \\
triple & 0.8 & \bfseries 0.400\pm0.001 & 0.399\pm0.001 & -0.2 & 667\pm  6 & \bfseries 446\pm  4 & 33.1 & \bfseries 
0.240\pm0.001 & 0.235\pm0.000 & -2.0 \\
node & 0.8 & \bfseries 0.440\pm0.002 & 0.431\pm0.001 & -2.0 & 520\pm  2 & \bfseries 380\pm  1 & 27.0 & \bfseries 
0.271\pm0.001 & 0.259\pm0.002 & -4.6 \\
relation & 0.4 & 0.413\pm0.023 & \bfseries 0.430\pm0.029 & 4.0 & 768\pm123 & \bfseries 343\pm 97 & 55.3 & 
0.249\pm0.025 & \bfseries 0.254\pm0.027 & 1.9 \\
relation & 0.6 & 0.437\pm0.014 & \bfseries 0.439\pm0.007 & 0.6 & 617\pm 42 & \bfseries 337\pm 64 & 45.4 & \bfseries
0.269\pm0.013 & 0.263\pm0.008 & -2.2 \\
relation & 0.8 & \bfseries 0.447\pm0.016 & 0.441\pm0.008 & -1.3 & 541\pm 83 & \bfseries 336\pm 67 & 37.8 & 
\bfseries 0.275\pm0.010 & 0.266\pm0.005 & -3.5 \\
\hline
 \multicolumn{2}{|c|}{Max boost (\%)}  &  &  & 16.7 &  &  & 65.8 &  &  & 15.5 \Tstrut\\
 \multicolumn{2}{|c|}{Mean boost (\%)}  &  &  & 3.2 &  &  & 44.4 &  &  & 1.1 \\
\hline
\end{tabular}
  }
  \caption{Synthetic scenario: performance comparison on KG completion task in both \underline{single} and \underline{linked} settings across various \underline{sampling} types and rates $p$. Each result is presented as a mean value and standard deviation. The superior performance for each metric, sampling and $p$ combination is highlighted. The \underline{boost} is the performance improvement the framework achieves over a single graph.  Please note that the results on relation sampling datasets should not be directly compared amongst themselves as the test was not preserved, causing high standard deviation (see \ref{sec:sampling}).}
  \label{tab:results-synthetic}

\end{table*}
\egroup

The following observation emerges when comparing the effects of different sampling strategies on boost. Triple sampling yielded a significantly higher boost compared to node sampling. Interestingly, node sampling maintained a higher base performance on the single graph than achieved with triple sampling ($Hits@10$ of $0.546$ vs $0.347$). This potentially explains the lower boost observed with node sampling. It indicates that while node sampling retains more integral information about the graph, triple sampling, being more disruptive, allows for a higher improvement using our framework. 

\begin{figure}[!htb]
    \centering
    \includegraphics[width=\linewidth]{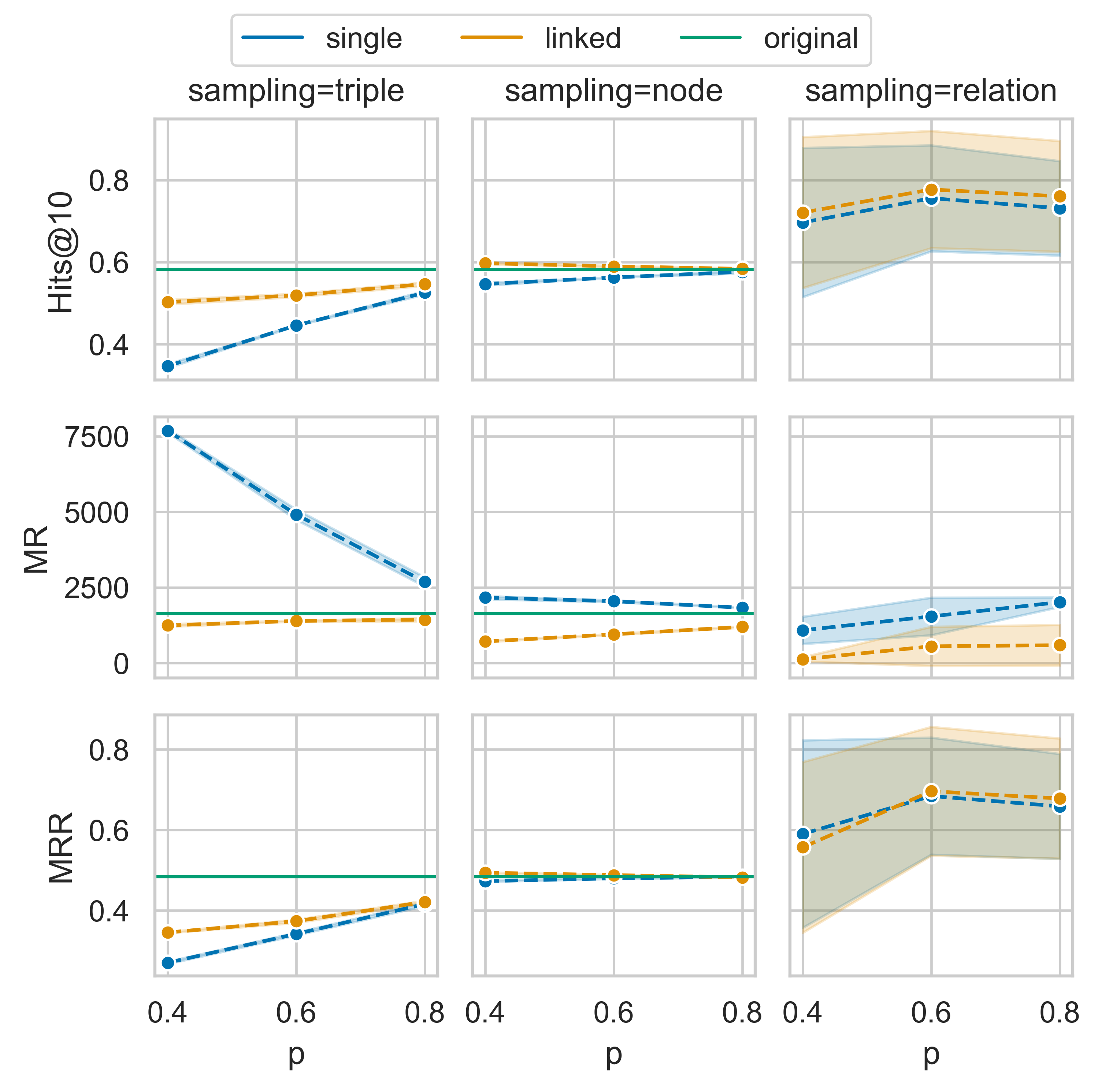}
    \caption{WN18RR: Performance across different sampling strategies and varying rates $p$. Two settings are shown: training on the \underline{single} and \underline{linked} graph. The \underline{green line} shows the performance on the original graph ($p=1.0$). The shaded areas represent the standard deviation across multiple runs. Please note that the relation sampling datasets should not be directly compared amongst themselves as the test was not preserved, causing high standard deviation (see Section \ref{sec:sampling}).}
    \label{fig:results_WN18RR}
\end{figure}

In Figure \ref{fig:results_WN18RR}, we can observe that while using node sampling, performance on a single graph does not decline significantly with lower $p$, and the potential for improvement is limited. Conversely, when sampling triples, there is a significant performance drop, and our framework could help to mitigate it. Therefore, increasing the number of nodes can lead to better outcomes than triples when expanding one's KG. Results for other datasets also confirm this statement (see Appendix \ref{sec:supplementary_results}: Figure \ref{fig:results_FB15K237}, \ref{fig:results_WD50K}).

\subsection{Real-world scenario}

Results from experiments in the real-world scenario are presented in Table \ref{tab:results-real}. We only showcased DKGs created by sampling triples, as it yielded the most significant differences in efficacy on the single graph (see Section \ref{sec:results-synthetic}). This choice allows a more convenient analysis of the framework's effectiveness. For each combination of (dataset, $p$), we recorded the efficacy of the trained model on a single graph and utilised the proposed framework by linking it to external KG. We explored two GKGs as potential external sources for each dataset, given that finding a suitable match is not straightforward. Similar to the previous scenario, the crucial metric remains the boost, which again shows dependency on the $p$ value. 

Under the most rigorous conditions ($40\%$ of triples), we observed the highest efficacy improvement, with a $14.0\%/-6.9\%/10.8\%$ $Hits@10$ boost on WN18RR-ConcepNet, FB15k-237-YAGO3-10, and WD50K-FB15k-237 pairs, respectively. A significant success was on WN18RR and WD50K. This suggests that our framework can enrich the embeddings of small-scale KGs in real-life situations. However, using FB15k-237 or less strict settings, the framework hurt performance. While our framework shows promise, performance improvement is not guaranteed in all situations. Further, in most cases, only one of two GKGs provided improvement. This confirms the mentioned requirement that the chosen GKG must be well-suited to the DKG.

\subsection{Common findings} \label{sec:common-findings}

In both scenarios, we observed varying degrees of improvement across different metrics. The highest boost was observed for MR, followed by Hits@10, with MRR registering the most minor enhancement. Notably, in some cases, the boost for the MR metric was negative, even while positive improvements were noted for other metrics. We found that determining the optimal loss was not obvious, as there was a slight variance in mean performance on the downstream task. Nevertheless, insights from the ablation study indicated that the trained models predict lower scores for incorrect linking triples. On average, the score of incorrect links was lower by $5.00\%$ while correct links' score only by $0.77\%$ (see Appendix \ref{sec:supplementary_results}: Figure \ref{fig:loss_comparison}). This indicates that the proposed loss can incorporate alignment probability into embeddings.

\bgroup

\begin{table*}[htb!]
  \centering
  \adjustbox{width=\linewidth}{%
    \sisetup{
      separate-uncertainty=true,
      multi-part-units=single,
      table-align-uncertainty=true,
      detect-weight=true,
      retain-zero-uncertainty=true
    }
    \setlength{\tabcolsep}{3pt}

    \begin{tabular}{|l
                S[table-format=1.1]||
                S[table-format=1.3(3)]|
                S[table-format=1.3]|
                S[table-format=1.3]|
                S[table-format=2.1]||
                S[table-format=4.0(3)]|
                S[table-format=4.0]|
                S[table-format=4.0]|
                S[table-format=1.1]||
                S[table-format=1.3(3)]|
                S[table-format=1.3]|
                S[table-format=1.3]|
                S[table-format=2.1]|}
    \hline
     & & \multicolumn{4}{c||}{$Hits@10\uparrow$} & \multicolumn{4}{c||}{$MR\downarrow$} & \multicolumn{4}{c|}{$MRR\uparrow$} \Tstrut\\
     {dataset} & {p}  & {single} & \multicolumn{2}{c|}{linked} & {boost(\%)} & {single} & \multicolumn{2}{c|}{linked}
    & {boost(\%)} & {single} & \multicolumn{2}{c|}{linked} & {boost(\%)} \\
    \hline\hline
     &   &  & {CN} & {FB} &  &  & {CN} & {FB} &  &  & {CN} & {FB} &  \Tstrut\\
    WN18RR & 0.4 & 0.347\pm0.005 & \bfseries 0.395 & 0.338 & 14.0 & 7681\pm 62 & \bfseries 1508 & 4050 & 80.4 & 
    \bfseries 0.270\pm0.003 & 0.259 & 0.262 & -3.2 \\
    WN18RR & 0.6 & 0.446\pm0.001 & \bfseries 0.471 & 0.435 & 5.7 & 4908\pm172 & \bfseries 963 & 2685 & 80.4 & \bfseries
    0.342\pm0.003 & 0.330 & 0.330 & -3.3 \\
    WN18RR & 0.8 & 0.525\pm0.004 & \bfseries 0.527 & 0.504 & 0.3 & 2685\pm155 & \bfseries 750 & 1725 & 72.1 & \bfseries
    0.416\pm0.005 & 0.390 & 0.400 & -3.8 \\
    \hline
    \multicolumn{2}{|c||}{Max boost (\%)} &  \multicolumn{3}{c}{}  & 14.0 & \multicolumn{3}{c}{} & 80.4 & \multicolumn{3}{c}{} & -3.2 \Tstrut\\
    \multicolumn{2}{|c||}{Mean boost (\%)} &  \multicolumn{3}{c}{}  & 6.7 & \multicolumn{3}{c}{} & 77.6 & \multicolumn{3}{c}{} & -3.5 \\ 
    \hline \hline
     &   &  & {CN} & {Y3-10} &  &  & {CN} & {Y3-10} &  &  & {CN} & {Y3-10} & \Tstrut\\
    FB15K237 & 0.4 & \bfseries 0.360\pm0.002 & 0.330 & 0.335 & -6.9 & 315\pm  4 & 315 & \bfseries 310 & 1.4 & \bfseries
    0.204\pm0.001 & 0.189 & 0.192 & -5.6 \\
    FB15K237 & 0.6 & \bfseries 0.398\pm0.001 & 0.359 & 0.362 & -9.1 & \bfseries 242\pm  1 & 254 & 251 & -3.9 & 
    \bfseries 0.227\pm0.001 & 0.202 & 0.208 & -8.6 \\
    FB15K237 & 0.8 & \bfseries 0.442\pm0.000 & 0.387 & 0.395 & -10.6 & \bfseries 194\pm  2 & 213 & 214 & -9.5 & 
    \bfseries 0.254\pm0.001 & 0.221 & 0.225 & -11.5 \\
    \hline
    \multicolumn{2}{|c||}{Max boost (\%)} &  \multicolumn{3}{c}{}  & -6.9 & \multicolumn{3}{c}{}  & 1.4 & \multicolumn{3}{c}{} & -5.6 \Tstrut\\
    \multicolumn{2}{|c||}{Mean boost (\%)} &  \multicolumn{3}{c}{} & -8.9 &  \multicolumn{3}{c}{} & -4.0 &  \multicolumn{3}{c}{} & -8.6 \\
    \hline  \hline
     &   &  & {FB} & {Y3-10} &  &  & {FB} & {Y3-10} &  &  & {FB} & {Y3-10} &  \Tstrut\\
    WD50K & 0.4 & 0.283\pm0.001 & \bfseries 0.313 & 0.284 & 10.8 & 1713\pm 42 & \bfseries 758 & 961 & 55.8 & 
    0.164\pm0.001 & \bfseries 0.181 & 0.163 & 10.6 \\
    WD50K & 0.6 & 0.353\pm0.000 & \bfseries 0.361 & 0.327 & 2.4 & 931\pm 20 & \bfseries 558 & 644 & 40.0 & 
    0.208\pm0.001 & \bfseries 0.212 & 0.190 & 2.0 \\
    WD50K & 0.8 & \bfseries 0.400\pm0.001 & 0.396 & 0.358 & -0.9 & 667\pm  6 & \bfseries 450 & 510 & 32.6 & \bfseries 
    0.240\pm0.001 & 0.236 & 0.210 & -1.8 \\
    \hline
     \multicolumn{2}{|c||}{Max boost (\%)} &  \multicolumn{3}{c}{}   & 10.8 &  \multicolumn{3}{c}{} & 55.8 & \multicolumn{3}{c}{} & 10.6\Tstrut \\
      \multicolumn{2}{|c||}{Mean boost (\%)}&  \multicolumn{3}{c}{}  & 4.1 &  \multicolumn{3}{c}{} & 42.8 &  \multicolumn{3}{c}{} & 3.6 \\
    \hline
    \end{tabular}
  
  }
  \caption{Real-world scenario: performance comparison on KG completion task in both \underline{single} and \underline{linked} settings across various datasets sampled using triple sampling and rates $p$. The DSKGs were linked to two of the following GKGs: ConceptNet (\underline{CN}), FB15k-237 (\underline{FB}) and YAGO3-10(\underline{Y3-10}). Each result is presented as a mean value and standard deviation, where applicable. The superior performance for each metric, dataset and $p$ combination is highlighted in bold. The \underline{boost} is the performance improvement of linking a GKG that yielded the higher score over training on a single graph.}
\label{tab:results-real}
\end{table*}
\egroup

\section{Limitations and Future Work}

One identified limitation is that the method may encounter challenges when DKG and GKG have a small number of common entities. A potential future direction involves addressing this issue by enhancing the method's robustness to the absence of an entity in GKG.

Moreover, in the scenario of small, domain-specific KGs already containing all necessary information for the completion task, our method—by adding general knowledge—might not substantially enhance performance. Nevertheless, it holds potential benefits for tasks requiring more distant connections than KG completion.

\section{Conclusion}

In this paper, we proposed the framework for enriching embeddings of domain-specific KG by aligning and linking entities to general-purpose KG. The framework is general and modular, meaning we can use any alignment method, representation model and loss function. We utilized a simple alignment method based on the textual embeddings of entities and their neighbourhood. Notably, this alignment approach did not require additional annotation, making it readily applicable. Moreover, we proposed the weighted loss function, which could help mitigate entity alignment's negative effects. 

We examined the proposed framework in extensive experiments, both in synthetic and real-world scenarios. The evaluation methodology was tailored to simulate the early stages of KG development, which are rigorous conditions where a KG may lack comprehensive information. Our results demonstrated that utilizing the knowledge from GKGs could significantly improve performance on the downstream task. However, it is essential to highlight that the degree of improvement depends on the particular domain-specific KG and linked general-purpose KG. This research underscores the possibilities of harnessing established GKGs to strengthen emerging KGs, offering a direction for improving their utility in real-world scenarios. Given that enriching small-scale KGs is an unexplored research area, this paper not only presents a unique approach but also signals a potential pathway for future research.

\section{Acknowledgments}

This work was funded by the European Union under the Horizon Europe grant OMINO – Overcoming Multilevel INformation Overload (grant number 101086321, \url{http://ominoproject.eu/}). Views and opinions expressed are those of the authors alone and do not necessarily reflect those of the European Union or the European Research Executive Agency. Neither the European Union nor the European Research Executive Agency can be held responsible for them. It was also co-financed with funds from the Polish Ministry of Education and Science under the programme entitled International Co-Financed Projects, grant no. 573977. This work was also funded in part by the National Science Centre, Poland under CHIST-ERA Open \& Re-usable Research Data \& Software  (grant number 2022/04/Y/ST6/00183) as well Department of Artificial Intelligence, Wroclaw Tech.

\section{Bibliographical References}\label{sec:reference}
\bibliographystyle{lrec-coling2024-natbib}
\bibliography{references}

\section{Language Resource References}
\label{lr:ref}
\bibliographystylelanguageresource{lrec-coling2024-natbib}
\bibliographylanguageresource{languageresource}

\FloatBarrier
\clearpage
\appendix

\section{Supplementary results} \label{sec:supplementary_results}

\begin{figure}[!htb]
    \centering
    \includegraphics[width=\linewidth]{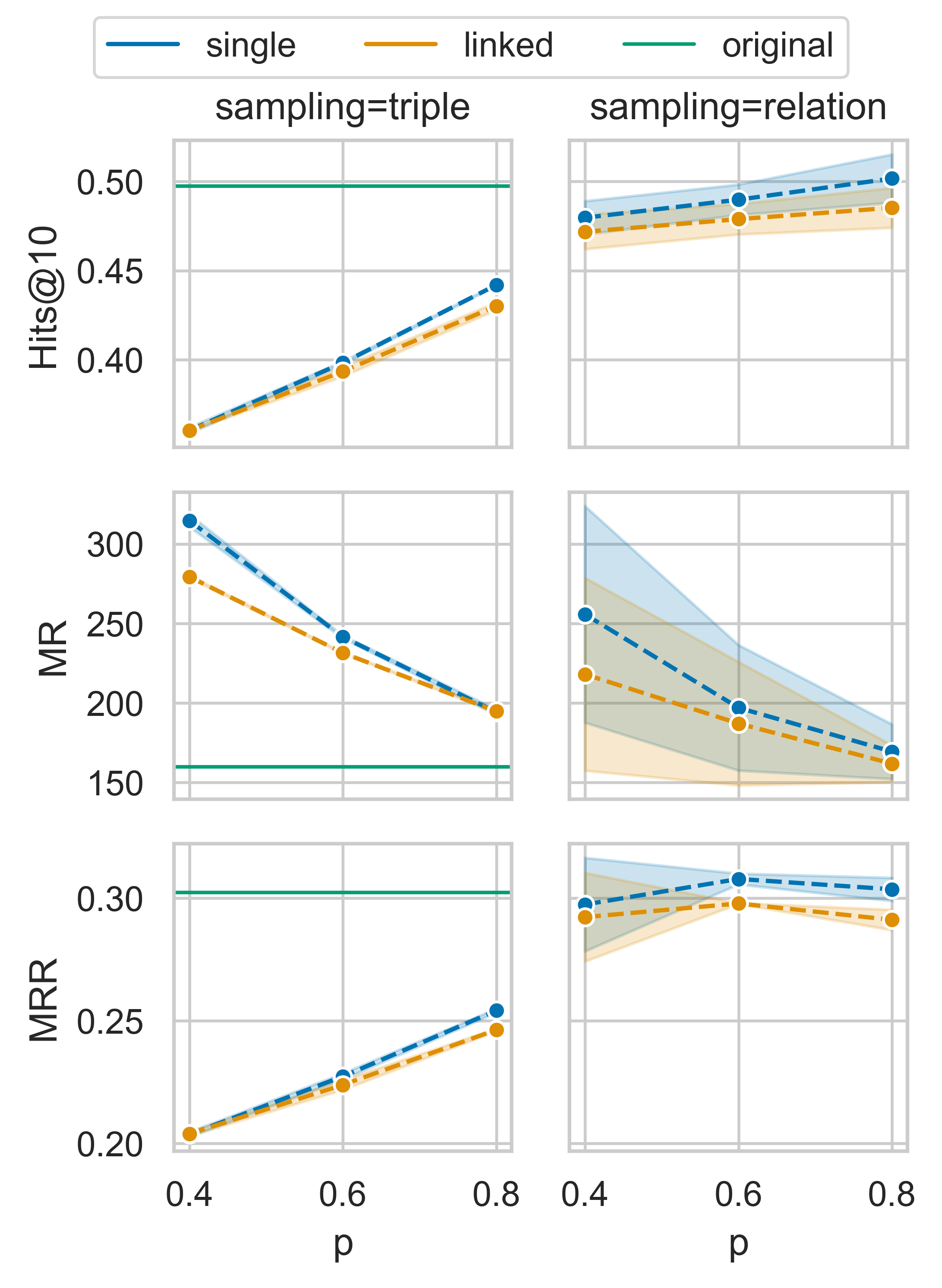}
    \caption{FB15k-237:  Performance across different sampling strategies and varying rates $p$. Two settings are shown: training on the \underline{single} and \underline{linked} graph. The \underline{green line} shows the performance on the original graph ($p=1.0$). The shaded areas represent the standard deviation across multiple runs. Please note that the relation sampling datasets should not be directly compared amongst themselves as the test was not preserved, causing high standard deviation (see Section \ref{sec:sampling}).}
    \label{fig:results_FB15K237}
\end{figure}

\begin{figure}[!htb]
    \centering
    \includegraphics[width=\linewidth]{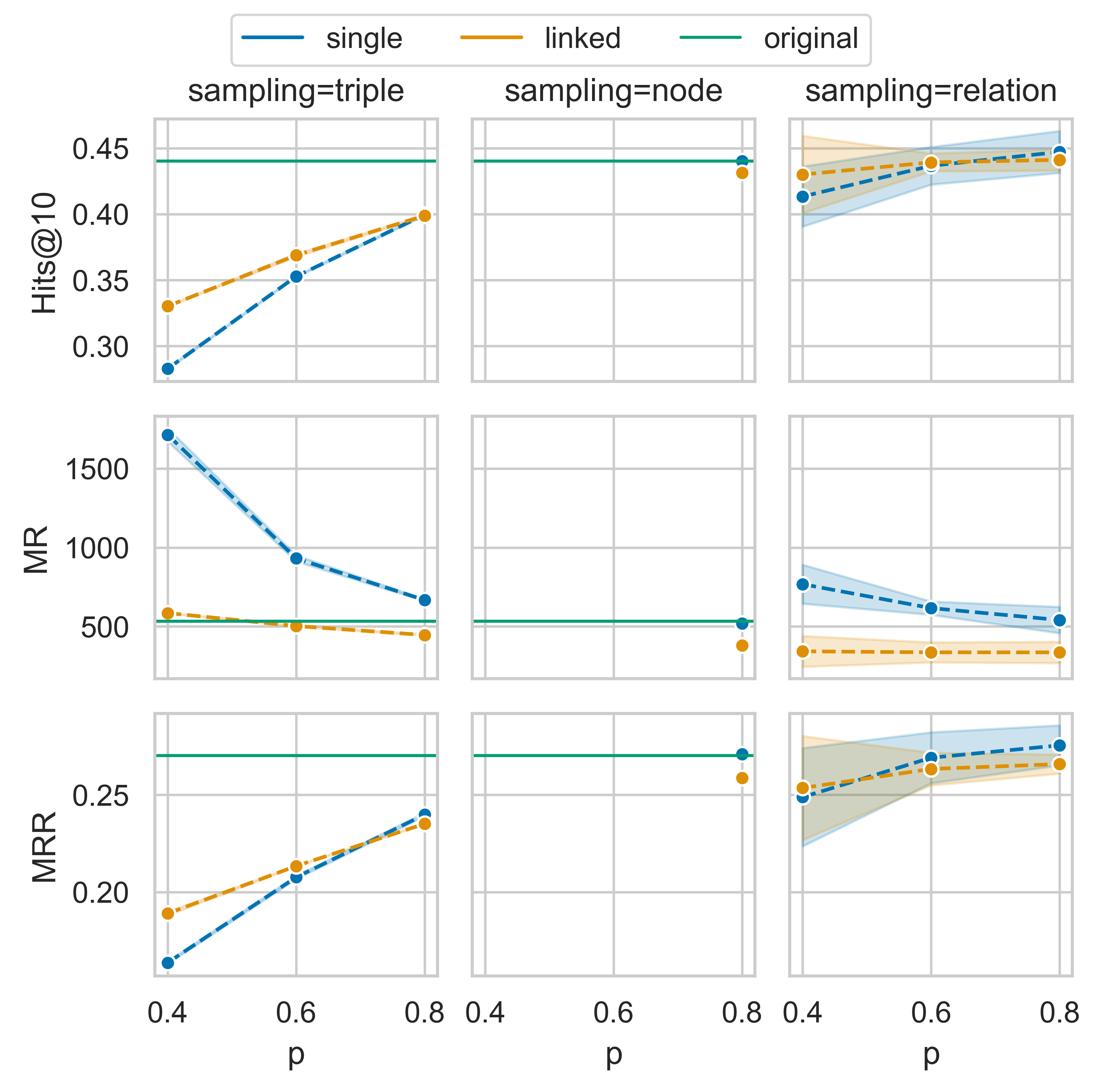}
    \caption{WD50K:  Performance across different sampling strategies and varying rates $p$. Two settings are shown: training on the \underline{single} and \underline{linked} graph. The \underline{green line} shows the performance on the original graph ($p=1.0$). The shaded areas represent the standard deviation across multiple runs. Please note that the relation sampling datasets should not be directly compared amongst themselves as the test was not preserved, causing high standard deviation (see Section \ref{sec:sampling}).} 
    \label{fig:results_WD50K}
\end{figure}

We extend the results of the synthetic scenario discussed in Section \ref{sec:results-synthetic} by providing performance analysis across different sampling strategies on Figures \ref{fig:results_FB15K237} and \ref{fig:results_WD50K} for datasets FB15k-237 and WD50K, respectively.

We further explore the common findings outlined in Section \ref{sec:common-findings} by conducting a comparative analysis of predicted scores for linking triples. This analysis contrasts models trained using the proposed loss function against those trained with the standard loss function.

\begin{figure*}[!htb]
    \centering
    \includegraphics[width=1.8\columnwidth]{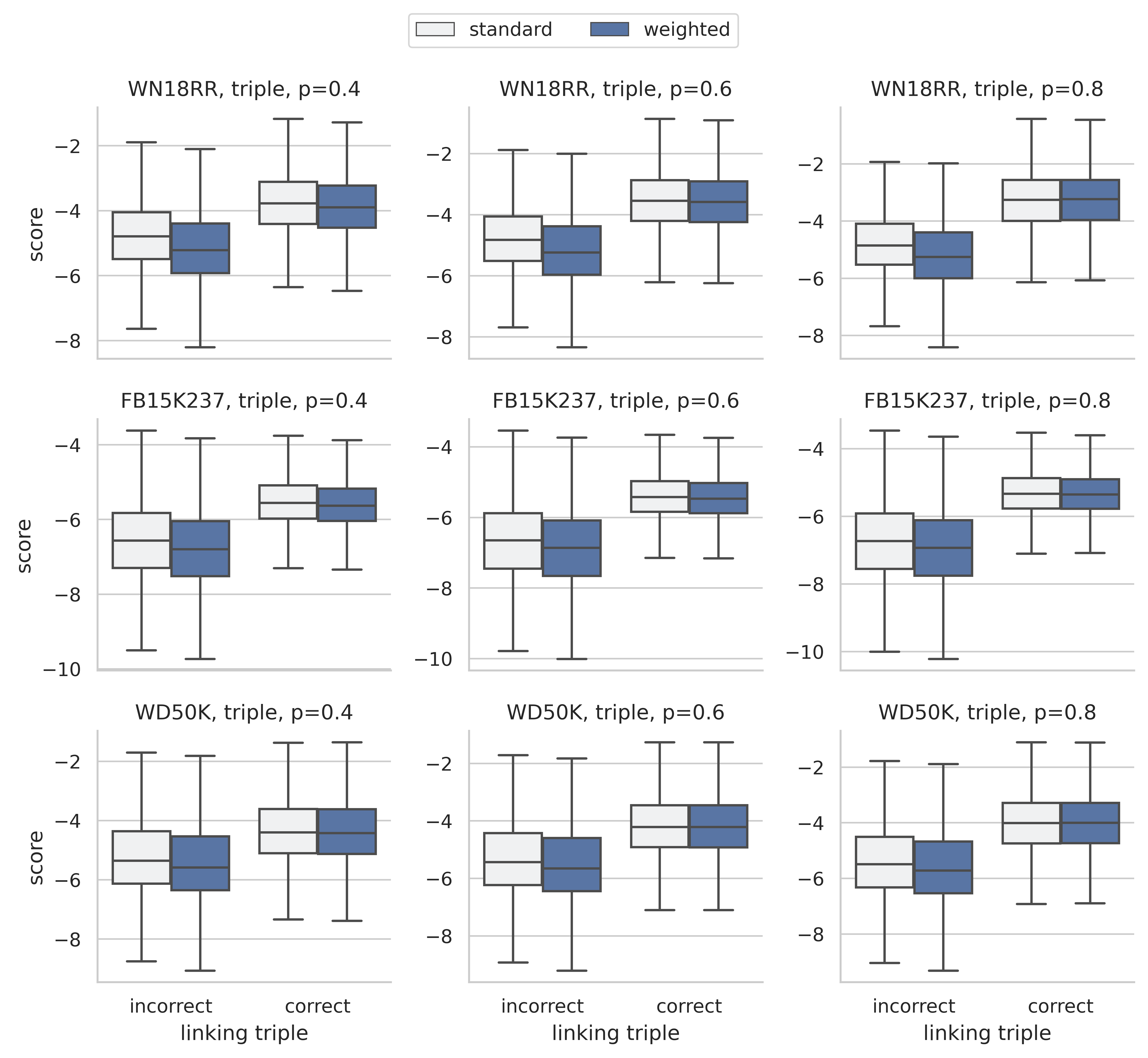}
    \caption{Comparative analysis of predicted scores for incorrect and correct linking triples from the training set for each setting (dataset, sampling, $p$).} 
    \label{fig:loss_comparison}
\end{figure*}

\section{Supplementary statistics of datasets} 
\label{sec:supplementary_stats}

We provide detailed statistics of the original and sampled datasets in Tables \ref{tab:original-stats} and \ref{tab:sampled-stats}, respectively.

\begin{table*}[!htbp]
  \centering
  \adjustbox{width=0.7\textwidth}{%
    \sisetup{
      separate-uncertainty=true,
      table-align-uncertainty=true,
      retain-zero-uncertainty=true
    }
    \setlength{\tabcolsep}{4pt}

\begin{tabular}{|l|S[table-format=7]
                S[table-format=5]
                S[table-format=5]|
                S[table-format=7]
                S[table-format=5]
                S[table-format=5]|
                S[table-format=3]
                S[table-format=3]
                S[table-format=3]|}
\hline
{} & \multicolumn{3}{c|}{Triples} & \multicolumn{3}{c|}{Entities} & \multicolumn{3}{c|}{Relations} \Tstrut \\
{dataset} & {train} & {val} & {test} & {train} & {val} & {test} & {train} & {val} & {test} \\
\hline
WN18RR* & 86835 & 2817 & 2923 & 40714 & 4835 & 4985 & 11 & 11 & 11 \Tstrut \\
FB15k-237*\dag & 272115 & 17526 & 20438 & 14505 & 9799 & 10317 & 237 & 237 & 237 \\
WD50K* & 164631 & 22429 & 45284 & 40107 & 16500 & 22732 & 473 & 297 & 347 \\
ConceptNet\dag & 3423004 & 0 & 0 & 1787373 & 0 & 0 & 47 & 47 & 47 \\
YAGO3-10\dag & 1079040 & 4978 & 4982 & 123143 & 7914 & 7906 & 37 & 37 & 37 \\
\hline
\end{tabular}
  }
  \caption{Statistics of the original datasets used in our experiments. The table shows the number of triples, entities, and relations across the training (\underline{train}), validation (\underline{val}), and \underline{test} sets for each dataset. Statistics refer to original datasets. Some were sampled or cropped during experiments (see Sections \ref{sec:sampling} and \ref{sec:model_setup}). Datasets used as DSKG are marked with *, while datasets used as GKG are marked with \dag.}
  \label{tab:original-stats}
\end{table*}

\begin{table*}[!htbp]
  \centering
  \adjustbox{width=\linewidth}{%
    \sisetup{
      separate-uncertainty=true,
      table-align-uncertainty=true,
      retain-zero-uncertainty=true
    }
    \setlength{\tabcolsep}{2pt}

\begin{tabular}{|l
                S[table-format=1.1]|
                S[table-format=6(5)]
                S[table-format=5(4)]
                S[table-format=5(4)]|
                S[table-format=5.0(5)]
                S[table-format=5.0(5)]
                S[table-format=5.0(5)]|
                S[table-format=3.0(1)]
                S[table-format=3.0(1)]
                S[table-format=3.0(1)]|}
                \hline
{} & {} & \multicolumn{3}{c|}{Triples} & \multicolumn{3}{c|}{Entities} & \multicolumn{3}{c|}{Relations} \Tstrut \\
{sampling} & {p} & {train} & {val} & {test} & {train} & {val} & {test} & {train} & {val} & {test} \\
\hline
\hline
\multicolumn{11}{|c|}{\textbf{WN18RR}} \Tstrut\\
\hline
triple & 0.4 & 34734(0) & 2817(0) & 2923(0) & 30975(57) & 4835(0) & 4985(0) & 11(0) & 11(0) & 11(0) \\
triple & 0.6 & 52101(0) & 2817(0) & 2923(0) & 36178(27) & 4835(0) & 4985(0) & 11(0) & 11(0) & 11(0) \\
triple & 0.8 & 69468(0) & 2817(0) & 2923(0) & 39093(28) & 4835(0) & 4985(0) & 11(0) & 11(0) & 11(0) \\
node & 0.4 & 28815(78) & 2817(0) & 2923(0) & 16286(0) & 4835(0) & 4985(0) & 11(0) & 11(0) & 11(0) \\
node & 0.6 & 46715(90) & 2817(0) & 2923(0) & 24428(0) & 4835(0) & 4985(0) & 11(0) & 11(0) & 11(0) \\
node & 0.8 & 66311(157) & 2817(0) & 2923(0) & 32571(0) & 4835(0) & 4985(0) & 11(0) & 11(0) & 11(0) \\
relation & 0.4 & 20367(18482) & 511(625) & 545(616) & 14485(10417) & 933(1168) & 989(1150) & 4(0) & 4(0) & 4(0) \\
relation & 0.6 & 57790(19777) & 1765(624) & 1827(635) & 32006(7832) & 3135(1000) & 3228(1025) & 7(0) & 7(0) & 7(0) 
\\
relation & 0.8 & 61586(17153) & 1852(598) & 1902(610) & 34809(4855) & 3286(949) & 3353(974) & 9(0) & 9(0) & 9(0) \\
\hline
\multicolumn{11}{|c|}{\textbf{FB15k-237}} \Tstrut\\ 
\hline
triple & 0.4 & 108846(0) & 17526(0) & 20438(0) & 14187(10) & 9799(0) & 10317(0) & 237(0) & 223(0) & 224(0) \\
triple & 0.6 & 163269(0) & 17526(0) & 20438(0) & 14352(6) & 9799(0) & 10317(0) & 237(0) & 223(0) & 224(0) \\
triple & 0.8 & 217692(0) & 17526(0) & 20438(0) & 14439(7) & 9799(0) & 10317(0) & 237(0) & 223(0) & 224(0) \\
relation & 0.4 & 103999(16860) & 6364(891) & 7419(1056) & 12884(220) & 5457(707) & 5981(722) & 95(0) & 87(2) & 
89(3) \\
relation & 0.6 & 156537(20322) & 9562(1011) & 11186(1268) & 13779(352) & 7218(571) & 7747(589) & 142(0) & 133(3) & 
134(3) \\
relation & 0.8 & 224750(25236) & 14309(1632) & 16748(1881) & 14332(75) & 8725(633) & 9291(618) & 190(0) & 178(2) & 
180(2) \\
\hline
\multicolumn{11}{|c|}{\textbf{WD50K}} \Tstrut\\
\hline
triple & 0.4 & 65852(0) & 22429(0) & 45284(0) & 34164(32) & 16500(0) & 22732(0) & 434(2) & 297(0) & 347(0) \\
triple & 0.6 & 98779(0) & 22429(0) & 45284(0) & 37091(41) & 16500(0) & 22732(0) & 453(6) & 297(0) & 347(0) \\
triple & 0.8 & 131705(0) & 22429(0) & 45284(0) & 38957(16) & 16500(0) & 22732(0) & 464(3) & 297(0) & 347(0) \\
node & 0.8 & 147774(42) & 22429(0) & 45284(0) & 32086(0) & 16500(0) & 22732(0) & 450(1) & 297(0) & 347(0) \\
relation & 0.4 & 62701(11210) & 7865(1591) & 16014(3259) & 25204(2662) & 7700(956) & 11535(1186) & 189(0) & 112(6) 
& 127(5) \\
relation & 0.6 & 88487(12893) & 11529(1838) & 23332(3923) & 29090(3143) & 10258(1183) & 14848(1580) & 284(0) & 
173(5) & 201(6) \\
relation & 0.8 & 121499(15858) & 16250(2317) & 32947(4969) & 33181(2970) & 12848(1161) & 18053(1489) & 378(0) & 
238(4) & 275(5) \\
\hline
\end{tabular}
  }
  \caption{Statistics of the sampled datasets used in our experiments. The table shows the number of triples, entities, and relations across the training (\underline{train}), validation (\underline{val}), and \underline{test} sets for each dataset.}
  \label{tab:sampled-stats}
\end{table*}

\section{Computational complexity} 

First, denote $n$ and $o$ as the number of DKG triples and entities, respectively. RotatE time complexity is $O(n)$, and space complexity is $O(o)$. Further, in our method we add GKG with $m$ triples and $p$ entities, linking each entity of DKG with $k$ nearest neighbors in GKG. Assuming the same embedding dimension, the time complexity of the proposed method is $O(n + m + k*o)$, with space complexity $O(o + p)$. 

\section{Alignment: entity representation}

Based on the preliminary results, we selected the best-performing method for entity representation, described in Section \ref{sec:alignment_linking}. In this section, we provide an ablation study.

The proposed concatenation method yielded promising results, achieving, on average, a $10\%$ higher accuracy during entity alignment compared to the absence of concatenation ($0.901$ vs. $0.819$). Without concatenation, we averaged embeddings across entity labels, outgoing neighbors' labels, and ingoing neighbors' labels. This experiment was conducted on all pairs of datasets utilized in the synthetic scenario that allows for such evaluation.

\end{document}